%% file: main.tex
\documentclass[10pt,twocolumn,letterpaper]{article}
\usepackage[pagenumbers]{cvpr}
\input{preamble}

\title{GaLF-ST: Gaussian Latent Fields\\for Spatial Transcriptomics Reconstruction}
\author{Baichuan Yang$^{1,2}$ \quad Yifan Luo$^{1}$ \quad Yong Liu$^{1}$ \quad Bing Su$^{1}$\\
{\normalsize $^{1}$Renmin University of China, Beijing \qquad $^{2}$Zhongguancun Academy}\\
{\small\ttfamily yang031119@outlook.com \quad 2024202715@ruc.edu.cn}\\
{\small\ttfamily liuyonggsai@ruc.edu.cn \quad bingsu@ruc.edu.cn}\\
{\small Corresponding author: Bing Su}}
\hypersetup{pdftitle={GaLF-ST: Gaussian Latent Fields for Spatial Transcriptomics Reconstruction},
            pdfauthor={Baichuan Yang, Yifan Luo, Yong Liu, Bing Su}}
\begin{document}
\raggedbottom
\maketitle
\begingroup\renewcommand{\thefootnote}{}\footnotetext{We will release the appendix and code with a future version of this preprint.}\endgroup
\input{sections/00_abstract}
\input{sections/01_introduction}
\input{sections/02_related_work}
\input{sections/03_method}
\FloatBarrier
\input{sections/04_experiments}
\input{sections/05_conclusion}
\input{sections/06_llm_usage}

{\small\bibliographystyle{ieeenat_fullname}\bibliography{references}}
\end{document}

%% file: preamble.tex
\usepackage{mathtools}
\usepackage{multirow}
\usepackage{array}
\usepackage{tabularx}
\usepackage{microtype}
\usepackage{placeins}
\definecolor{cvprblue}{rgb}{0.21,0.49,0.74}
\usepackage[pagebackref,breaklinks,colorlinks,allcolors=cvprblue]{hyperref}
\def\confName{CVPR}
\def\confYear{2027}
\newcommand{\method}{GaLF-ST}
\newcommand{\R}{\mathbb{R}}

\newcommand{\x}{\mathbf{x}}
\newcommand{\y}{\mathbf{y}}
\newcommand{\z}{\mathbf{z}}
\newcommand{\cvec}{\mathbf{c}}

\newcommand{\softplus}{\operatorname{softplus}}

\newcommand{\Fig}[1]{Fig.~\ref{#1}}
\newcommand{\Tab}[1]{Tab.~\ref{#1}}

%% file: sections/00_abstract.tex
\begin{abstract}
Spatial transcriptomics (ST) measures gene expression at known tissue locations to reveal tissue organization and cellular function. Expression prediction requires modeling both local spatial relationships and associations among genes. Existing methods typically model expression using coordinate networks without explicit spatial parameterization. To address limited spatial sampling and sparse expression observations of ST, some methods firstly fit intermediate embeddings and then train expression predictors, but a latent representation suited to expression reconstruction may be difficult to predict from spatial coordinates. The two training stages can impose different demands on the latent representation. In this paper, we propose GaLF-ST, a Gaussian latent field method in which learnable Gaussian centers and anisotropic covariances determine the weights of latent vectors at each location. A shared nonlinear decoder maps their weighted average to multigene expression. In GaLF-ST, observed expression error jointly updates the Gaussian geometry, latent vectors, and decoder parameters. We further establish a unified benchmark across seven dataset groups and three held-out fractions, comparing GaLF-ST with four representative methods using the same locations, splits, gene panels, and scoring rules. Across tissues and observation densities, GaLF-ST achieves leading overall reconstruction performance, recovering spatial gene expression patterns and preserving tissue domains from limited observations while substantially reducing runtime and GPU memory usage.
\end{abstract}

%% file: sections/01_introduction.tex
\begin{figure*}[t]
\centering
\includegraphics[width=\textwidth]{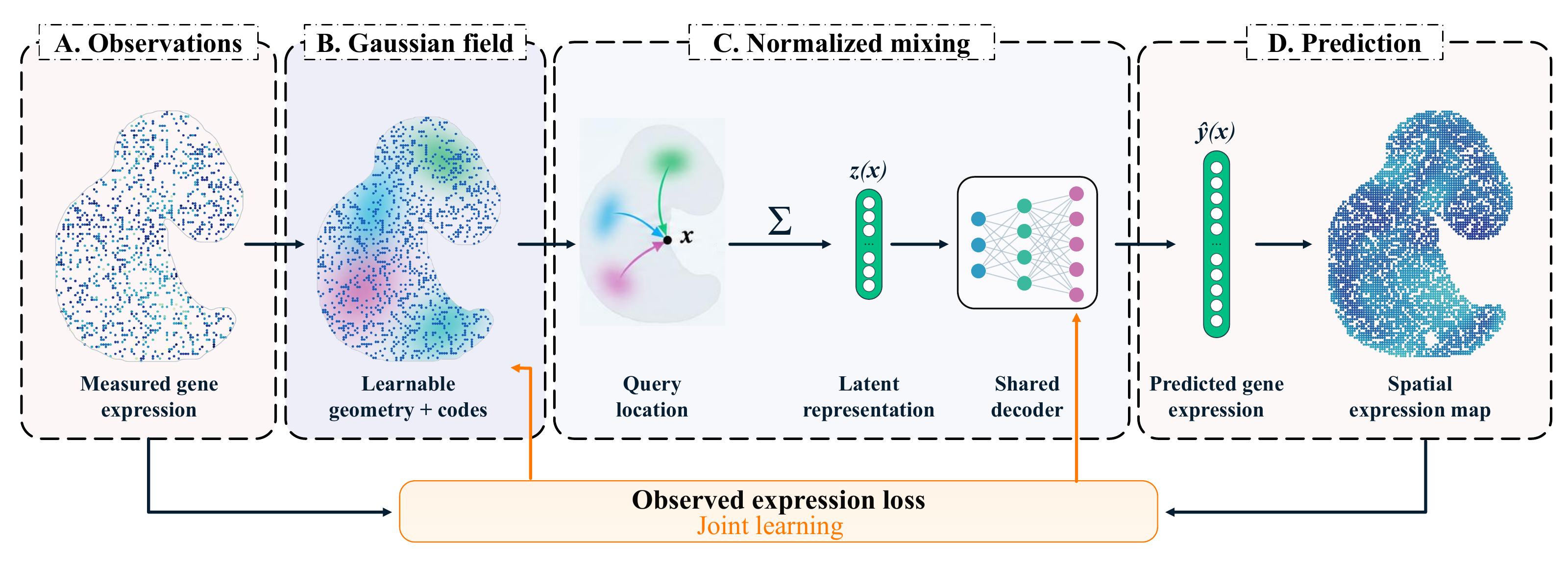}
\caption{Overview of \method. (A) Observed gene expression. (B) Learnable Gaussian geometry and latent coefficients. (C) At query location $\x$, normalized Gaussian responses combine the coefficients into $\z(\x)$. A shared decoder maps this latent state to predicted expression. (D) Predictions at held-out locations. The loss on observed expression jointly updates the geometry, coefficients, and decoder.}
\label{fig:method}
\end{figure*}
\section{Introduction}
\label{sec:intro}
Spatial transcriptomics (ST) measures gene expression at tissue locations to characterize tissue organization and cell states~\cite{stahl}. Limited spatial sampling leaves some expression variation unobserved~\cite{slideseq,hdst}. Predicting multigene expression at unmeasured locations yields denser tissue maps. These maps support tissue-domain identification, marker-gene localization, and comparisons of regional expression patterns~\cite{stage,suica,cps}. Reconstruction must capture both gradual changes within tissue regions and rapid changes at their boundaries. It must also model thousands of sparsely observed genes jointly~\cite{spark,trendsceek,suica}.

Coordinate neural networks are an established approach to modeling expression as a continuous function of spatial coordinates. STAGE aligns expression latent variables with coordinates, while STINR combines implicit neural representations with reference expression priors~\cite{stage,stinr}. These designs improve prediction from limited observations. However, local spatial relationships remain implicit in shared network weights. Without explicit parameters for local position, scale, and orientation, the networks represent tissue regions and boundaries indirectly.

The high dimensionality and sparsity of expression data also motivate auxiliary representations. SUICA first constructs dense graph autoencoder embeddings and trains a coordinate network to predict them. It then freezes that network and fine-tunes the decoder~\cite{suica}. CPS uses a teacher to extract neighborhood expression states. During student training, it freezes the shared decoder to preserve the teacher's latent space~\cite{cps}. Both methods fit intermediate embeddings in separate stages. Latent states learned for expression reconstruction may be difficult to infer from spatial location. Once frozen, the coordinate network or decoder cannot respond to prediction error in that stage.

Reliable comparison also requires a unified benchmark. ST datasets vary in tissue type, developmental stage, sampling platform, section size, observation density, and expression sparsity. Existing studies evaluate selected tissues and sections with different gene panels, masking procedures, and aggregation rules. These differences limit comparisons across studies~\cite{stage,stinr,suica,cps}.

We propose \method, a Gaussian latent field method, and establish a unified benchmark for spatial prediction. Inspired by explicit Gaussian representations~\cite{gaussians,gaussianimage}, \method{} uses spatial basis functions with learnable centers and covariances. These parameters determine each basis function's contribution at different tissue locations. The normalized weights~\cite{nrbf} define a weighted average of learnable, low-dimensional latent vectors. Expression prediction error jointly trains the spatial representation, latent states, and decoder in a unified procedure. A shared nonlinear decoder predicts all genes jointly from the resulting vector at every location. Observed expression error jointly updates the Gaussian geometry, latent vectors, and decoder parameters from initialization (\Fig{fig:method}). Our unified benchmark covers diverse tissues and observation densities using common locations, splits, gene panels, and scoring rules. It compares \method{} with four representative methods across seven dataset groups and three held-out fractions. Across these 21 settings, \method{} achieves the best overall reconstruction performance, demonstrating its ability to recover expression patterns and tissue structure from limited observations.

Our main contributions are threefold.
\begin{itemize}
  \item We propose a Gaussian latent field method that explicitly models and predicts gene expression using Gaussian primitives with learnable centers and anisotropic covariances, improving prediction quality while substantially reducing time and computational costs.
  \item We jointly optimize Gaussian geometry, low-dimensional latent coefficients, and a shared decoder using a common loss on observed expression. Spatial prediction and compact multigene representations are learned together.
  \item We establish a unified benchmark that evaluates expression reconstruction and tissue-domain preservation under fair comparison conditions, demonstrating the overall reconstruction advantage of \method.
\end{itemize}

%% file: sections/02_related_work.tex
\section{Related Work}
\label{sec:related}
\subsection{Spatial Gene Expression Reconstruction}
Most methods for spatial expression reconstruction model gene expression using coordinate networks. STAGE~\cite{stage} aligns autoencoder latent variables with spatial coordinates, and STINR~\cite{stinr} directly regresses expression. SUICA~\cite{suica} predicts graph-autoencoder embeddings. CPS~\cite{cps} distills multiscale context from observed neighborhoods into a coordinate network.

\paragraph{Global frequency and local scale.}
These methods map continuous coordinates to expression or latent features at different spatial scales. Representing high-frequency detail is therefore a key design choice. SUICA uses SIREN periodic activations. CPS augments its student INR with Fourier frequencies sampled from a Gaussian distribution. This sampling encodes coordinates; it does not construct learnable spatial Gaussian primitives. Periodic activations~\cite{siren} and Fourier features~\cite{fourierfeatures} can reduce the low-frequency bias of coordinate multilayer perceptrons~\cite{spectralbias}. However, global hyperparameters still set the frequency range. Kriging and Gaussian-process regression~\cite{gpml} offer another class of spatial models. spaVAE~\cite{spavae} also uses Gaussian-process latent variables. Their commonly used stationary covariances describe tissue with shared length scales.

\paragraph{Implicit structure and local adjustment.}
SUICA and CPS reduce output dimensionality through embedding prediction, but local positions, extents, and orientations remain implicit in shared network weights.

\paragraph{Staged optimization and the expression objective.}
STAGE and STINR train end to end, whereas SUICA and CPS fit embeddings in separate stages. SUICA freezes its coordinate mapping after warm-up and then refines the decoder. CPS freezes the decoder during student updates. Neither method continuously optimizes the query mapping and decoder together under the final expression objective.

\subsection{Explicit Gaussian Representations}
Existing explicit Gaussian representations commonly model visual signals with learnable Gaussian primitives. 3D Gaussian Splatting~\cite{gaussians} jointly optimizes their positions, anisotropic covariances, opacities, and view-dependent colors. It projects the primitives onto the image plane and applies depth-ordered alpha blending for real-time novel-view synthesis. GaussianImage~\cite{gaussianimage} adapts this approach to 2D image representation and compression. It replaces depth sorting and alpha blending with a weighted sum of 2D Gaussian colors. Inspired by this adaptive spatial representation, \method{} constructs a Gaussian latent field over tissue coordinates.

%% file: sections/03_method.tex
\section{GaLF-ST}
\label{sec:method}
\subsection{Preliminaries}
\label{sec:preliminary}
Spatial transcriptomics (ST) measures gene expression and records where it occurs in tissue. After expression quantification and spatial matching, the data comprise an expression matrix and spatial coordinates for its rows~\cite{stinr}. Each row records transcript counts across genes at one sampled location; each column corresponds to a gene. A sampled location may contain multiple cells rather than a single cell~\cite{stinr}. Placing one gene's expression values at their corresponding coordinates produces its spatial expression map.

Sampling locations form regular arrays or irregular distributions, depending on the platform. ST data can thus be represented as spatial point sets with high-dimensional expression vectors~\cite{suica}. Zero entries may indicate absent expression or failed transcript capture. Spatial reconstruction must account for this sparse, high-dimensional expression and discrete sampling~\cite{suica}. Individual sections provide 2D locations. Multiple sections with shared spatial geometry can form 3D data~\cite{stage}.

We predict multigene expression at query locations from measured expression. Let $N$ be the number of observations and $d\in\{2,3\}$ the spatial dimension. The indices $i\in\{1,\ldots,N\}$ and $g\in\{1,\ldots,G\}$ identify locations and genes, respectively. We write the observations as $\mathcal O=\{(\x_i,\y_i)\}_{i=1}^{N}$. Here, $\x_i\in\R^d$ is a spatial coordinate, and $\y_i\in\R_+^G$ contains library-size-normalized, log-transformed expression for $G$ genes. The entry $y_{ig}$ denotes expression of gene $g$ at location $i$. The query set $\mathcal Q\subset\R^d$ contains coordinates only. Our goal is to learn $f:\R^d\rightarrow\R_+^G$ and complete the expression map at these locations. Appendix~B details expression preprocessing and the fixed coordinate normalization.

\subsection{Method}
\label{sec:method_design}
\subsubsection{Overview}
\label{sec:method_overview}
\Fig{fig:method} shows the four parts of \method. We place $K$ Gaussian primitives in tissue coordinate space, each with $q$ latent coefficients. At coordinate $\x\in\R^d$, normalized Gaussian responses combine these coefficients into a latent state $\z(\x)\in\R^q$. A shared decoder maps this state to multigene expression. Evaluating the model at query coordinates produces a spatial expression map.

During training, we pass fitting coordinates through the field and decoder and compare predictions with measured expression. This error jointly updates the geometry, coefficients, and decoder through the bottom feedback path in the figure. Internal validation selects the number of training epochs. We then refit the model on all observations. Prediction uses the same mixing and decoding operations at each query coordinate.

\subsubsection{Gaussian latent field}
\label{sec:gaussian_field}
Our design draws on the explicit Gaussian representations in 3D Gaussian Splatting~\cite{gaussians} and GaussianImage~\cite{gaussianimage}. Learnable centers and covariances control the location, extent, and direction of spatial sharing (\Fig{fig:method}B). Each primitive $k\in\{1,\ldots,K\}$ has a center $\boldsymbol\mu_k\in\R^d$, a covariance matrix $\boldsymbol{\Sigma}_k\in\R^{d\times d}$, and a learnable coefficient vector $\cvec_k\in\R^q$. The covariance explicitly describes spatial extent and orientation. The low-dimensional latent representation serves three purposes. First, shared latent factors capture dependencies among genes. Second, compression extracts shared expression structure to reduce the impact of ST zero inflation on reconstruction. Third, when $q\ll G$, the number of learnable expression coefficients falls from $KG$ to $Kq$. This reduces parameter overhead and improves model efficiency.

For primitive $k$ at coordinate $\x$, let $\ell_k(\x)$ be the Gaussian log-response and $w_k(\x)$ the mixing weight. We define these quantities and the latent state as
\begin{equation}
\begin{aligned}
 \ell_k(\x)&=-\tfrac12(\x-\boldsymbol\mu_k)^\top\boldsymbol{\Sigma}_k^{-1}(\x-\boldsymbol\mu_k),\\
 w_k(\x)&=[\operatorname{softmax}(\boldsymbol\ell(\x))]_k,\\
 \z(\x)&=\sum_{k=1}^{K}w_k(\x)\cvec_k.
\end{aligned}
\label{eq:field}
\end{equation}
The vector $\boldsymbol\ell(\x)=(\ell_1(\x),\ldots,\ell_K(\x))^\top$ collects all log-responses. Softmax normalizes them over the $K$ primitives. The normalized radial basis weights~\cite{nrbf,rbf} satisfy
\begin{equation}
    \sum_{k=1}^{K} w_k(\x) = 1.
\end{equation}
The latent state therefore depends on relative responses and lies in the convex hull of the coefficient vectors. For numerical stability, we subtract the largest input before computing softmax. This leaves the normalized weights unchanged~\cite{softmax}.

We parameterize each spatial covariance using a Cholesky factor. Let $\mathbf{L}_k\in\R^{d\times d}$ be lower triangular and set $\boldsymbol{\Sigma}_k=\mathbf{L}_k\mathbf{L}_k^\top$. The indices $a,b\in\{1,\ldots,d\}$ identify coordinate axes. Let $s_{ka}>0$ denote the $a$th diagonal scale and $\gamma_{kab}\in\R$ the free shear parameters for $a>b$. The factor entries are
\begin{equation}
 (\mathbf{L}_k)_{ab}=\begin{cases}
 s_{ka},&a=b,\\
 s_{ka}\tanh\gamma_{kab},&a>b,\\
 0,&a<b.
 \end{cases}
 \label{eq:covariance}
\end{equation}
Positive diagonal scales ensure that $\boldsymbol{\Sigma}_k$ is positive definite. There is one shear parameter in 2D and three in 3D. Let $\mathbf s_k=(s_{k1},\ldots,s_{kd})^\top$ be the scale vector, $\mathbf s_k^0$ its initial value, and $\boldsymbol\mu_k^0$ the initial center. The vector $\boldsymbol\eta_k\in\R^d$ parameterizes center displacement in units of the initial scales. The vector $\mathbf u_k\in\R^d$ controls scale changes, and $r>1$ sets their maximum multiplicative range. We define
\begin{equation}
\begin{aligned}
 \boldsymbol\mu_k&=\boldsymbol\mu_k^0+\mathbf s_k^0\odot\boldsymbol\eta_k,\\
 \mathbf s_k&=\mathbf s_k^0\odot\exp\!\bigl((\log r)\tanh\mathbf u_k\bigr).
\end{aligned}
\label{eq:geometry}
\end{equation}
Centers can move freely. Each scale remains between $1/r$ and $r$ times its initial value, and normalized shear remains between $-1$ and $1$.

\subsubsection{Shared expression decoding}
\label{sec:expression_decoder}
Following nonlinear approaches to low-dimensional expression modeling~\cite{scvi}, we use a shared decoder to map latent states to gene expression (\Fig{fig:method}C). Let $\mathcal F\subseteq\{1,\ldots,N\}$ be the nonempty set of fitting indices. For gene $g$, the fitting mean $m_g$ and population standard deviation $\sigma_g$ are
\begin{equation}
 m_g=\frac{\sum_{i\in\mathcal F}y_{ig}}{|\mathcal F|},\qquad
 \sigma_g=\sqrt{\frac{\sum_{i\in\mathcal F}(y_{ig}-m_g)^2}{|\mathcal F|}}.
 \label{eq:gene_statistics}
\end{equation}
Let $\mathcal V=\{g\in\{1,\ldots,G\}:\sigma_g>0\}$ index genes with nonzero variance in the fitting subset. The decoder $D_\theta:\R^q\rightarrow\R_+^G$ has hidden width $H$. Its parameters $\theta$ comprise weights $\mathbf{W}_1\in\R^{H\times q}$ and $\mathbf{W}_2\in\R^{G\times H}$, and biases $\mathbf b_1\in\R^H$ and $\mathbf b_2\in\R^G$. For an input latent state $\z\in\R^q$, the output preactivation vector $\mathbf o_\theta(\z)\in\R^G$ and gene predictions are
\begin{equation}
\begin{aligned}
 \mathbf o_\theta(\z)&=\mathbf{W}_2\operatorname{GELU}(\mathbf{W}_1\z+\mathbf b_1)+\mathbf b_2,\\
 [D_\theta(\z)]_g&=\begin{cases}
 \sigma_g\softplus([\mathbf o_\theta(\z)]_g),&g\in\mathcal V,\\
 m_g,&g\notin\mathcal V.
 \end{cases}
\end{aligned}
\label{eq:prediction}
\end{equation}
The hidden layer uses GELU~\cite{gelu}. Softplus ensures nonnegative predictions, and $\sigma_g$ restores each gene's expression scale. Constant genes are predicted at $m_g$. We initialize the final layer to produce values near the fitting means. Appendix~B.1 gives the initialization details.

\subsubsection{Joint optimization with expression error}
\label{sec:joint_optimization}
Let $\mathcal B\subseteq\mathcal F$ be a nonempty mini-batch of observation indices. The prediction for gene $g$ at location $i$ is $\hat y_{ig}=[D_\theta(\z(\x_i))]_g$. We measure error relative to observed variation using the standardized expression loss
\begin{equation}
 \mathcal L_{\rm std}=\frac{1}{|\mathcal B||\mathcal V|}
 \sum_{i\in\mathcal B}\sum_{g\in\mathcal V}
 \frac{(\hat y_{ig}-y_{ig})^2}{\sigma_g^2},
 \label{eq:loss}
\end{equation}
Training minimizes $\mathcal L_{\rm std}$ using each positive standard deviation directly. If $\mathcal V$ is empty, the loss is zero. Backpropagation jointly updates $\theta$, $\cvec_k$, $\boldsymbol\eta_k$, $\mathbf u_k$, and $\gamma_{kab}$. Spatial sharing and expression prediction are thus learned together. At each epoch, we shuffle all fitting indices and process them in mini-batches, including the final partial batch.

We reserve observations indexed by $\mathcal I$ for validation; the remaining indices form $\mathcal F$. Let $t$ count completed training epochs. We compute validation loss $v_t$ using Eq.~\eqref{eq:loss} on $\mathcal I$ with statistics from $\mathcal F$. Validation runs at initialization ($t=0$) and after each epoch. We initialize the reference loss to $v_0$ and the patience counter to zero. An improvement is effective when the loss strictly decreases and its relative reduction from the reference reaches $\delta$. This resets the counter and updates the reference. Otherwise, the counter increases by one. Training stops when the counter reaches $P$ or the epoch count reaches $T_{\max}$. If training stops at epoch $T$, we select the earliest epoch $t^\star$ attaining the minimum $v_t$ over $t=0,\ldots,T$. This selection is independent of $\delta$. We then recompute statistics and initialization on all observations and refit for $t^\star$ epochs from the original seed (Appendix~B.1).

\subsubsection{Expression prediction at query locations}
\label{sec:query_prediction}
We fix the refitted field, decoder, and statistics. At each normalized query coordinate $\x^\ast\in\mathcal Q$, the predicted expression vector $\hat\y(\x^\ast)\in\R_+^G$ is
\begin{equation}
 \hat\y(\x^\ast)=D_\theta\!\left(\sum_{k=1}^{K}w_k(\x^\ast)\cvec_k\right).
 \label{eq:query_prediction}
\end{equation}
Evaluating all query locations along the B--C--D path in \Fig{fig:method} produces the multigene expression map.

%% file: sections/04_experiments.tex
\section{Experiments}
\label{sec:experiments}
\begin{figure*}[!t]
  \centering
  \includegraphics[width=0.95\textwidth]{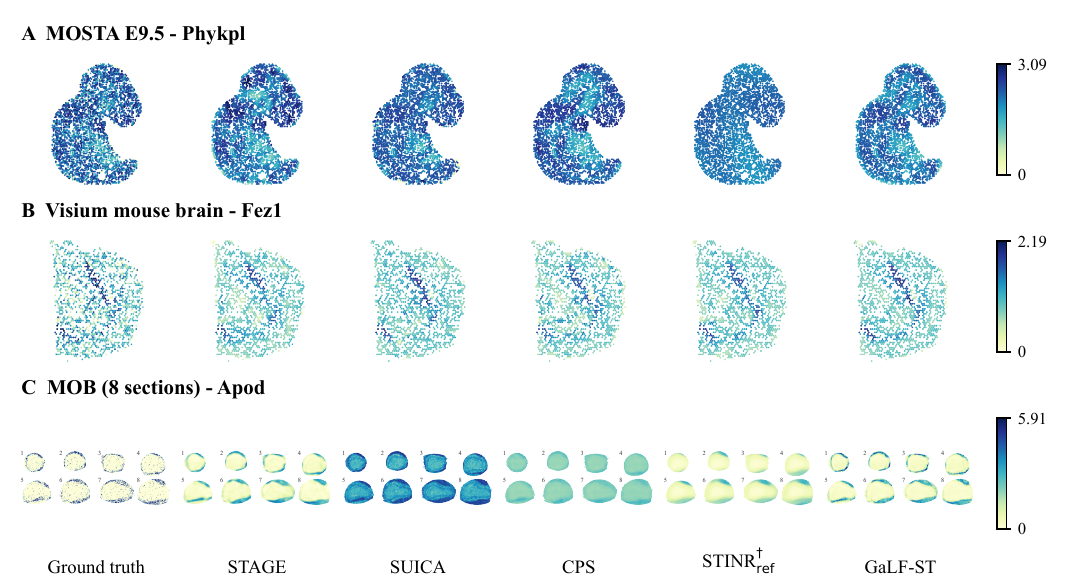}
  \caption{Spatial expression patterns predicted by \method{} in three datasets. All hidden locations are shown on a shared linear color scale, capped at the 99th percentile of nonzero observed expression.}
  \label{fig:selected_genes}
\end{figure*}
\input{tables/table_01}

\subsection{Datasets and unified evaluation}
We evaluate reconstruction as observations become sparser, recovery of spatial expression patterns, and preservation of tissue structure. The datasets cover whole embryos, brain regions sampled at near-cellular resolution, human cortex, and a 3D organ stack. Appendix~B summarizes their sizes and expression sparsity. MOSTA E9.5 and E16.5~\cite{mosta} test whole-embryo reconstruction at early and late stages of organogenesis. Visium mouse brain tests reconstruction with larger sampling spots. Slide-seqV2 hippocampus~\cite{slideseqv2} tests local brain reconstruction from irregularly distributed beads. DLPFC~\cite{dlpfc} contains 12 human cortical sections from three donors. MOB extends evaluation to 3D coordinates from 8 aligned sections. We fit each dataset independently. MOSTA E10.5, Visium mouse brain, and Slide-seqV2 hippocampus provide additional validation; full results appear in Appendix~I.

\paragraph{Comparison setup.}
We hold out measured locations and fit each model using expression at the remaining locations. Predictions are compared with the held-out measurements at three observation densities. Appendix~C.8 includes comparisons with global means, neighborhood means, and distance-weighted interpolation.
STAGE~\cite{stage}, SUICA~\cite{suica}, CPS~\cite{cps}, STINR~\cite{stinr}, and \method{} use the same locations, held-out splits, and gene panels. We hold out 20\%, 50\%, or 80\% of locations, with nested observation sets and three random-split repeats. Each task uses 3,000 genes selected only from the common 20\% observed core. All expression undergoes the same library-size normalization and log transformation. Held-out expression and its library sizes are excluded from training. Checkpoints are selected using only internal validation on observed data. For MOB, we hold out points within each section and retain the remaining observations. This evaluates 3D point completion in the aligned stack. Appendix~H reports results for entirely held-out sections.

\method{} uses the same model configuration across datasets. Baselines retain their public architectures and training procedures within the shared task setup. CPS also uses raw counts from the same observed points. Its queries contain coordinates only, and output calibration uses observed points only. The MOB implementation extends CPS to 3D. STINR$_{\rm ref}^{\dagger}$ uses additional reference information from cortex for DLPFC and olfactory bulb for the other datasets. All available results enter the numerical comparison. Bold marks the best means at full precision. We retrain SUICA with its native linear expression targets and convert predictions to the common log scale for scoring. Appendix~B details the settings, reference sources, count calibration, and implementation adaptations.

\paragraph{Metrics.}
The main table reports mean squared error (MSE) on the normalized log scale and PCC$_g$ for each gene's spatial pattern. Spot-wise PCC$_s$ measures agreement between centered expression profiles, while Cos$_s$ measures their uncentered direction. Cos$_f$ is computed after flattening the matrix of held-out locations by genes; it is not the mean of spot-wise cosines. Similarity and expression error provide complementary information. Flattened scores also reflect differences in abundance between genes and should be interpreted alongside gene-wise spatial metrics.

Appendix~I also reports mean absolute error (MAE), normalized mean squared error (nMSE), normalized root mean squared error (nRMSE), rank correlations, clustering diagnostics, and SDs. nMSE and nRMSE normalize by observed gene variability to reduce the influence of gene-scale differences on MSE. nMSE emphasizes large standardized errors, whereas nRMSE takes the square root of each gene's MSE, normalizes it, and then averages across genes to measure error magnitude relative to observed variability.

To assess preservation of tissue domains, we use the adjusted Rand index (ARI) and normalized mutual information (NMI)~\cite{ari,nmi} to compare expression-derived clusters with tissue annotations. ARI measures chance-corrected agreement in pairwise group assignments, while NMI measures the information shared by the two partitions. Higher values indicate better agreement, complementing expression error with an assessment of tissue structure. All scores use the fixed gene panel. Undefined correlations and cosines with zero norms contribute zero. For DLPFC, we average repeats within sections and then weight the 12 sections equally. For MOB, we pool held-out points.

\begin{figure*}[!t]
  \centering
  \includegraphics[width=\textwidth]{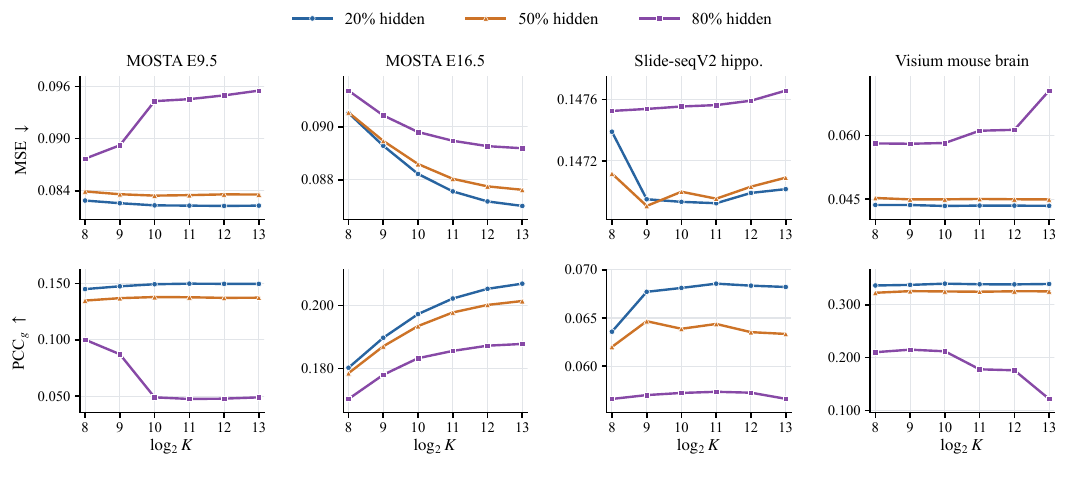}
  \caption{Gaussian primitive-count ablation with uniform center initialization and fixed $q=32$. Detailed results are in Appendix~G.3.}
  \label{fig:capacity_latent}
\end{figure*}

\paragraph{Model initialization and optimization settings.}
We implement \method{} in PyTorch~\cite{pytorch}. The default configuration uses $K=1024$ Gaussian primitives, latent dimension $q=32$, and decoder hidden width $H=128$. Model parameters are fitted independently for each dataset. To cover the coordinate bounding box, we repeatedly bisect its current widest cell axis until it contains $K$ Cartesian cells. Their midpoints initialize the Gaussian centers. Each center inherits scales based on sampling density at its nearest fitting location. Center displacements, scale adjustments, and shear parameters start at zero. Coefficients are independent Gaussian draws with mean zero and standard deviation $0.01$. AdamW~\cite{adamw} optimizes all parameters with learning rate $10^{-3}$. Internal validation uses 20\% of observations. The stopping parameters are $P=20$, $\delta=10^{-3}$ (i.e., $0.1\%$), and $T_{\max}=1000$. Further details appear in Appendix~B.1.

\subsection{Gene expression prediction}
\label{sec:results}
\subsubsection{Overview}
The seven dataset groups and three held-out fractions yield 21 settings. Compared with STAGE, \method{} improves MSE, nRMSE, PCC$_g$, and PCC$_s$ in 20, 21, 21, and 20 settings, respectively. It outperforms SUICA on all four metrics in every setting. \Tab{tab:unified_reconstruction} shows representative results, with complete results in Appendix~I. Compared with CPS, \method{} improves these metrics in 19, 19, 17, and 16 settings. MSE reductions range from 0.9\% to 44.4\% in the 19 favorable settings. On Visium, MSE is 0.4\% higher at 20\% holdout and 1.8\% higher at 80\% holdout. In these experiments, the default configuration of \method{} reduces peak GPU memory usage by 59.24\%--99.80\% and total training and prediction time by 56.38\%--99.51\% compared with the other methods. Detailed measurements are provided in Appendix~C.9.

\subsubsection{Effect of observation density}
Higher held-out fractions leave fewer observed locations and lower observation density. Reconstruction performance declines more gradually for \method{} as observations become sparser. STAGE and SUICA show increasing expression error and decreasing spot-wise correlation, while \method{} better preserves both aspects of reconstruction quality. Compared with STINR$_{\rm ref}^{\dagger}$, \method{} improves MSE, PCC$_g$, and PCC$_s$ at all held-out fractions on DLPFC and E16.5, and at 20\% and 50\% holdout on E9.5. The reference method performs better at 80\% holdout on E9.5. Spatial correlation on E9.5 declines appreciably as the held-out fraction increases, indicating a loss of recoverable spatial information in the smaller section. E16.5 shows a smaller decline in spatial correlation and little change in error, consistent with its larger absolute number of retained observations.

\subsubsection{Point completion in a stacked organ}
MOB tests the representation on an aligned organ stack with the sparsest expression in this collection. \method, STAGE, and STINR$_{\rm ref}^{\dagger}$ with its tissue-matched reference have similar MSE. However, \method{} consistently achieves higher PCC$_g$ (\Tab{tab:unified_reconstruction}). In highly sparse data, small differences in full-matrix error can accompany larger differences in spatial patterns. As the held-out fraction increases, expression error changes little while spatial correlation declines. These results concern point completion using one model fitted jointly to the aligned sections.

\subsection{Preservation of tissue structure}
\label{sec:biology}
\input{tables/biology_domains}
We compare clusters derived from reconstructed expression with tissue annotations on MOSTA and DLPFC. We reduce expression profiles to 30 dimensions using PCA, then apply KMeans with seed 0 and 10 initializations. Labels only determine the cluster count and provide reference partitions for ARI/NMI. \Tab{tab:biology_domains} reports two embryo stages and human cortex. On DLPFC, \method{} achieves higher ARI and NMI than all four methods from related work at every held-out fraction. At 80\% holdout, ARI and NMI are approximately 15.1\% and 12.9\% higher than STAGE, respectively. The corresponding gains over CPS are 45.1\% and 38.1\%. Reconstructed expression thus preserves partitions aligned with cortical annotations under sparse observation. Among the embryo sections, the advantage is more pronounced for late-stage E16.5, where NMI remains highest at all three held-out fractions. Results are mixed for the early-stage embryo, with CPS competitive at higher held-out fractions. Appendix~C.7 provides additional experiments and metrics.

\subsection{Ablation studies}
\label{sec:controls}
\subsubsection{Explicit Gaussian geometry versus neural spatial mappings}
\label{sec:representation_ablation}
We test whether explicit spatial mixing improves reconstruction over a coordinate mapping encoded in network weights. We replace the Gaussian coordinate-to-latent field with a Fourier feature network (FFN) or SIREN. The decoder, inputs, expression supervision, joint training, and validation selection remain fixed. Network widths match the spatial module's parameter budget. This control tests the spatial representation within a common prediction pipeline.

The Gaussian field achieves lower MSE than FFN and SIREN in 11/21 and 19/21 settings, respectively. It achieves higher PCC$_g$ in 16/21 and 18/21 settings. The explicit Gaussian field therefore better preserves spatial correlation in most settings. Appendix~G.6 provides the experimental settings and detailed results.

\subsubsection{Gaussian primitive count}
\label{sec:capacity_ablation}
With $q=32$ fixed, we evaluate six primitive counts from $K=256$ to 8192 (\Fig{fig:capacity_latent}). Internal training and refitting on all observations each retain exactly $K$ primitives. Increasing $K$ does not consistently improve performance across datasets and held-out fractions, while spatial-field computation grows approximately linearly with $K$ for a fixed query count and latent dimension. At lower observed fractions, larger values of $K$ tend to overfit. Balancing computational cost, empirical performance, and overfitting risk, we recommend $K=1024$. In practice, $K$ can be adjusted according to section size and the density of observed locations.

\subsubsection{Latent dimension}
\label{sec:q_ablation}
With $K=1024$ fixed, we vary $q$ from 8 to 6000 and set decoder width to $H=\max(128,q)$. We average relative changes with equal weight across the 21 settings. Relative to $q=32$, $q=8$ increases MSE by 1.38\% and decreases PCC$_g$ by 3.73\%. Using $q=6000$ decreases MSE by 0.44\% and increases PCC$_g$ by 3.57\%. However, $q=6000$ has worse MSE and PCC$_g$ in 15/21 and 16/21 settings, respectively. Its gains are concentrated in a subset of settings. We retain the compact $q=32$ configuration. Appendix~G.5 provides curves and full results.

\subsubsection{Anisotropic kernels}
\label{sec:anisotropy_ablation}
With $K=1024,q=32$ fixed, we replace anisotropic kernels with learnable isotropic kernels. The geometric mean of the initial axis scales preserves the initial kernel volume. Anisotropy yields lower MSE, higher PCC$_g$, and lower MAE in 19/21, 20/21, and 19/21 settings, respectively. Removing anisotropy increases mean expression error and reduces spatial correlation. These modest but consistent gains support learnable orientations and axis scales for directional variation in tissue expression.

\subsubsection{Kernel center initialization}
\label{sec:init_ablation}
With $K=1024,q=32$ fixed, we replace deterministic cell centers with centers sampled uniformly at random within the same bounding box. The remaining training protocol is unchanged. Averaged equally across 21 settings, relative MSE and MAE increase by 0.0066\% and 0.0222\%, while PCC$_g$ decreases by 0.6743\%. In every setting, the magnitude of the relative MSE and MAE changes is below 0.48\%. Reconstruction is robust to these two initialization schemes. Full results appear in Appendix~G.4.

We also evaluate patience windows $P\in\{5,10,15,20,25,30\}$. Appendix~G.2 reports the results and training costs.

%% file: tables/table_01.tex
\ifdefined\GaLFCombinedTable\else\begin{table*}[t]\fi
\centering
\footnotesize
\setlength{\tabcolsep}{1.5pt}
\caption{Expression reconstruction results. Means over three seeds; DLPFC weights 12 sections equally, and MOB pools held-out points from eight sections. Bold indicates the best mean; $\dagger$ additional reference information. SDs are in Appendix~I.}
\label{tab:unified_reconstruction}
\begin{tabular*}{\textwidth}{@{\extracolsep{\fill}}lrrrrrrrrrrrrrrr@{}}
\toprule
Method & \multicolumn{5}{c}{20\% hidden} & \multicolumn{5}{c}{50\% hidden} & \multicolumn{5}{c}{80\% hidden} \\
\cmidrule(lr){2-6}\cmidrule(lr){7-11}\cmidrule(lr){12-16}
 & MSE $\downarrow$ & PCC$_g$ $\uparrow$ & PCC$_s$ $\uparrow$ & Cos$_f$ $\uparrow$ & Cos$_s$ $\uparrow$ & MSE $\downarrow$ & PCC$_g$ $\uparrow$ & PCC$_s$ $\uparrow$ & Cos$_f$ $\uparrow$ & Cos$_s$ $\uparrow$ & MSE $\downarrow$ & PCC$_g$ $\uparrow$ & PCC$_s$ $\uparrow$ & Cos$_f$ $\uparrow$ & Cos$_s$ $\uparrow$ \\
\midrule
\multicolumn{16}{l}{\textit{MOSTA E9.5}} \\
STAGE & 0.0987 & 0.0392 & 0.7364 & 0.7596 & 0.7604 & 0.1071 & 0.0332 & 0.7158 & 0.7389 & 0.7407 & 0.1230 & 0.0286 & 0.6811 & 0.7057 & 0.7073 \\
SUICA & 0.1337 & 0.0186 & 0.6455 & 0.6539 & 0.6539 & 0.1334 & 0.0164 & 0.6473 & 0.6542 & 0.6548 & 0.1420 & 0.0105 & 0.6239 & 0.6279 & 0.6326 \\
CPS & 0.0863 & 0.1260 & \textbf{0.7745} & 0.7962 & 0.7995 & 0.0881 & 0.1205 & 0.7692 & 0.7918 & 0.7950 & 0.0972 & \textbf{0.0885} & \textbf{0.7447} & 0.7701 & \textbf{0.7739} \\
STINR$_{\rm ref}^{\dagger}$ & 0.0922 & 0.0699 & 0.7469 & 0.7763 & 0.7764 & 0.0921 & 0.0681 & 0.7465 & 0.7759 & 0.7759 & \textbf{0.0935} & 0.0605 & 0.7421 & \textbf{0.7720} & 0.7721 \\
\textbf{\method} & \textbf{0.0823} & \textbf{0.1494} & 0.7744 & \textbf{0.8031} & \textbf{0.8005} & \textbf{0.0834} & \textbf{0.1381} & \textbf{0.7713} & \textbf{0.7996} & \textbf{0.7976} & 0.0943 & 0.0491 & 0.7400 & 0.7696 & 0.7704 \\
\midrule
\multicolumn{16}{l}{\textit{MOSTA E16.5}} \\
STAGE & 0.0972 & 0.0638 & 0.7056 & 0.7176 & 0.7172 & 0.0981 & 0.0604 & 0.7028 & 0.7145 & 0.7141 & 0.1002 & 0.0541 & 0.6958 & 0.7072 & 0.7069 \\
SUICA & 0.1296 & 0.0172 & 0.5996 & 0.6023 & 0.6044 & 0.1290 & 0.0162 & 0.6025 & 0.6050 & 0.6070 & 0.1300 & 0.0141 & 0.5974 & 0.5995 & 0.6016 \\
CPS & 0.1000 & 0.1575 & 0.7127 & 0.7214 & 0.7256 & 0.0998 & 0.1560 & 0.7130 & 0.7216 & 0.7260 & 0.1018 & 0.1553 & 0.7100 & 0.7187 & 0.7229 \\
STINR$_{\rm ref}^{\dagger}$ & 0.1009 & 0.1077 & 0.6878 & 0.7046 & 0.7049 & 0.1010 & 0.1072 & 0.6874 & 0.7042 & 0.7045 & 0.1011 & 0.1066 & 0.6873 & 0.7040 & 0.7044 \\
\textbf{\method} & \textbf{0.0878} & \textbf{0.1973} & \textbf{0.7334} & \textbf{0.7496} & \textbf{0.7478} & \textbf{0.0882} & \textbf{0.1935} & \textbf{0.7319} & \textbf{0.7481} & \textbf{0.7463} & \textbf{0.0898} & \textbf{0.1833} & \textbf{0.7266} & \textbf{0.7430} & \textbf{0.7414} \\
\midrule
\multicolumn{16}{l}{\textit{Human DLPFC}} \\
STAGE & 0.1357 & 0.0206 & 0.6140 & 0.6358 & 0.6365 & 0.1416 & 0.0177 & 0.5984 & 0.6193 & 0.6208 & 0.1600 & 0.0128 & 0.5565 & 0.5748 & 0.5779 \\
SUICA & 0.1679 & 0.0118 & 0.5463 & 0.5589 & 0.5633 & 0.1690 & 0.0110 & 0.5422 & 0.5548 & 0.5594 & 0.1757 & 0.0087 & 0.5228 & 0.5348 & 0.5403 \\
CPS & 0.1333 & 0.0412 & 0.6424 & 0.6631 & 0.6648 & 0.1352 & 0.0337 & 0.6352 & 0.6580 & 0.6599 & 0.1429 & 0.0192 & 0.6088 & 0.6354 & 0.6383 \\
STINR$_{\rm ref}^{\dagger}$ & 0.1229 & 0.0422 & 0.6497 & 0.6750 & 0.6760 & 0.1246 & 0.0370 & 0.6438 & 0.6691 & 0.6706 & 0.1327 & 0.0259 & 0.6181 & 0.6440 & 0.6468 \\
\textbf{\method} & \textbf{0.1216} & \textbf{0.0544} & \textbf{0.6534} & \textbf{0.6790} & \textbf{0.6794} & \textbf{0.1221} & \textbf{0.0477} & \textbf{0.6516} & \textbf{0.6772} & \textbf{0.6778} & \textbf{0.1261} & \textbf{0.0311} & \textbf{0.6403} & \textbf{0.6638} & \textbf{0.6675} \\
\midrule
\multicolumn{16}{l}{\textit{Mouse olfactory bulb}} \\
STAGE & 0.1409 & 0.0082 & 0.4520 & 0.4560 & 0.4585 & 0.1409 & 0.0088 & 0.4524 & 0.4567 & 0.4592 & 0.1424 & 0.0061 & 0.4450 & 0.4482 & 0.4512 \\
SUICA & 0.8810 & 0.0038 & 0.2091 & 0.2347 & 0.2325 & 0.9123 & 0.0040 & 0.2054 & 0.2322 & 0.2295 & 0.9826 & 0.0041 & 0.1998 & 0.2271 & 0.2249 \\
CPS & 0.2446 & 0.0317 & 0.3757 & 0.3732 & 0.3754 & 0.2464 & 0.0299 & 0.3689 & 0.3687 & 0.3682 & 0.2493 & 0.0310 & 0.3736 & 0.3709 & 0.3728 \\
STINR$_{\rm ref}^{\dagger}$ & 0.1406 & 0.0171 & 0.4498 & 0.4578 & 0.4611 & 0.1401 & 0.0226 & 0.4537 & 0.4619 & 0.4648 & 0.1400 & 0.0222 & 0.4543 & 0.4627 & 0.4655 \\
\textbf{\method} & \textbf{0.1377} & \textbf{0.0481} & \textbf{0.4650} & \textbf{0.4756} & \textbf{0.4761} & \textbf{0.1380} & \textbf{0.0462} & \textbf{0.4639} & \textbf{0.4743} & \textbf{0.4750} & \textbf{0.1386} & \textbf{0.0409} & \textbf{0.4616} & \textbf{0.4715} & \textbf{0.4727} \\
\bottomrule
\end{tabular*}
\ifdefined\GaLFCombinedTable\else\end{table*}\fi

%% file: tables/biology_domains.tex
\begin{table*}[t]
\centering\small
\setlength{\tabcolsep}{3pt}
\caption{Tissue-domain preservation measured by ARI/NMI. Labels are used only for evaluation. Bold marks the best mean; $\dagger$ denotes additional reference information. SDs appear in Appendix~C.7.}
\label{tab:biology_domains}
\begin{tabular*}{\textwidth}{@{\extracolsep{\fill}}llrrrrr@{}}
\toprule
Dataset & Hidden & STAGE & SUICA & CPS & STINR$_{\rm ref}^{\dagger}$ & \method \\
\midrule
MOSTA E9.5 & 20\% & 0.2526/0.4758 & 0.2590/0.4514 & 0.2963/0.5323 & 0.2137/0.4585 & \textbf{0.3082}/\textbf{0.5344} \\
MOSTA E9.5 & 50\% & 0.2209/0.4477 & 0.2601/0.4320 & \textbf{0.3036}/\textbf{0.5333} & 0.2229/0.4532 & 0.2905/0.5205 \\
MOSTA E9.5 & 80\% & 0.2392/0.4517 & 0.1596/0.2912 & \textbf{0.2730}/\textbf{0.4640} & 0.2071/0.4307 & 0.2681/0.3929 \\
\midrule
MOSTA E16.5 & 20\% & 0.4493/0.6018 & 0.4626/0.5689 & 0.4624/0.6060 & 0.2350/0.4607 & \textbf{0.4754}/\textbf{0.6307} \\
MOSTA E16.5 & 50\% & 0.4280/0.5967 & 0.4502/0.5555 & \textbf{0.4722}/0.6068 & 0.2291/0.4585 & 0.4601/\textbf{0.6167} \\
MOSTA E16.5 & 80\% & 0.4296/0.5869 & 0.4091/0.5244 & \textbf{0.4679}/0.6038 & 0.2318/0.4536 & 0.4372/\textbf{0.6040} \\
\midrule
DLPFC & 20\% & 0.2993/0.4629 & 0.2871/0.4353 & 0.3148/0.4717 & 0.2824/0.4428 & \textbf{0.3211}/\textbf{0.4889} \\
DLPFC & 50\% & 0.2926/0.4578 & 0.2617/0.4051 & 0.2929/0.4432 & 0.2516/0.4035 & \textbf{0.3454}/\textbf{0.4935} \\
DLPFC & 80\% & 0.2756/0.4183 & 0.1981/0.3058 & 0.2186/0.3421 & 0.1801/0.3233 & \textbf{0.3172}/\textbf{0.4724} \\
\bottomrule
\end{tabular*}
\end{table*}

%% file: sections/05_conclusion.tex
\begin{samepage}
\section{Conclusion}
\method{} represents sparse spatial transcriptomic data as a continuous field with directly optimizable local geometry. Learnable Gaussian centers and covariances determine how latent vectors are averaged at each location. A shared decoder maps the resulting vector to multigene expression. Expression prediction error jointly updates the geometry, latent vectors, and decoder. We also establish a unified evaluation protocol across seven dataset groups and three held-out fractions, covering 2D tissues and 3D stacks. Checkpoint selection uses observed data only. Across these settings, \method{} achieves strong overall reconstruction performance. It introduces a new Gaussian modeling paradigm for high-density spatial transcriptomics. Future work will extend the framework to 3D tissue reconstruction and multimodal integration to support the study of complex biological systems.
\par
\end{samepage}

%% file: sections/06_llm_usage.tex
\section*{LLM Usage Statement}
We used OpenAI Codex in preparing this work. For code and implementation, Codex generated project code that the authors reviewed, edited, or approved during interactive use (Generate). In the other reported areas, Codex refined author-produced content and supported author-directed work (Assist). These areas included manuscript writing and editing, related work and citations, and experiment design, execution, and analysis. They also included figures and visual materials, data-related work, and research ideas, hypotheses, claims, and conclusions. In particular, Codex helped run experiments and edit and format LaTeX. The authors remain responsible for the code, experiments, scientific claims, and final manuscript.